\documentclass[10pt, a4paper]{article}

\usepackage[final]{lrec2026}
\usepackage{booktabs}
\usepackage{multirow}
\usepackage{graphicx}
\usepackage{xcolor}
\usepackage{amsmath}
\usepackage{tabularx}
\usepackage{url}
\usepackage{hyperref}
\usepackage{xspace}
\usepackage{booktabs}
\usepackage{enumitem}

\usepackage{todonotes}

\newcommand{\quechua}{Quechua\xspace}
\newcommand{\spanish}{Spanish\xspace}
\newcommand{\constitution}{Peruvian Constitution\xspace}

\title{Giving Voice to the Constitution: Low-Resource Text-to-Speech for Quechua and Spanish Using a Bilingual Legal Corpus}

\name{John E. Ortega$^{1}$ \quad Rodolfo Zevallos$^{2}$ \quad Fabrício Carraro$^{3}$}

\address{$^{1}$Northeastern University, USA \\ $^{2}$Universitat Pompeu Fabre, Spain \\ $^{3}$Barcelona Supercomputing Center, Spain \\
     j.ortega@northeastern.edu,
     rodolfojoel.zevallos@upf.edu, fabricio.carraro@bsc.es\\}

\abstract{
We present a unified pipeline for synthesizing high-quality \quechua and \spanish speech for the Peruvian Constitution using three state-of-the-art text-to-speech (TTS) architectures: \emph{XTTS v2}, \emph{F5-TTS}, and \emph{DiFlow-TTS}. Our models are trained on independent Spanish and Quechua speech datasets with heterogeneous sizes and recording conditions, and leverage bilingual and multilingual TTS capabilities to improve synthesis quality in both languages. By exploiting cross-lingual transfer, our framework mitigates data scarcity in Quechua while preserving naturalness in Spanish. We release trained checkpoints, inference code, and synthesized audio for each constitutional article, providing a reusable resource for speech technologies in indigenous and multilingual contexts. This work contributes to the development of inclusive TTS systems for political and legal content in low-resource settings.
}
\begin{document}
\maketitleabstract

\section{Introduction}
Indigenous Andean communities in South America often face barriers where crucial information, such as laws and other political issues, is only communicated in the official high-resource language of the government (\spanish). One indigenous community found in Peru is a prime example of this notion in which the government intended to address barriers by translating its constitution from \spanish to \quechua, the largest indigenous language in Peru. For most Peruvians, the \constitution\ is central to civic life and is made officially available online\footnote{\url{https://www.congreso.gob.pe/biblioteca/constituciones_peru}}\footnote{\url{https://www.wipo.int/wipolex/en/legislation/details/21225}}\footnote{\url{https://biblioteka.sejm.gov.pl/konstytucje-swiata-peru/?lang=en}}\footnote{\url{https://www.constituteproject.org/constitution/Peru_2021}}; public portals\footnote{\url{https://biblioteka.sejm.gov.pl/konstytucje-swiata-peru/?lang=en}} and legal repositories provide the authoritative \spanish text and cataloged \quechua versions and translations. \cite{bewes1920new}

In order to facilitate further work in the political arena for Quechua, the authors of this paper target two principal goals: (1) \textbf{accessibility}---produce clear audio renditions of the \constitution\ in \quechua and \spanish for radio, legal aid, screen readers, and civic outreach; and (2) \textbf{reusability}---publish aligned text and synthesized speech that others can adopt for ASR/ST training, evaluation, and augmentation in Quechua-focused research.

This short paper is submitted to the \textbf{PoliticalNLP}\footnote{\url{https://sites.google.com/view/politicalnlp2026}} workshop at LREC~2026 and follows the conference formatting and anonymization guidelines. The proposed resource is lightweight enough to be developed by a student team, yet sufficiently broad to support practical experimentation in \quechua and \spanish speech technologies.

Section~\ref{sec:related} reviews relevant prior work. Section~\ref{sec:legal} introduces the role of TTS in low-resource political and legal domains. Our modeling framework and training strategy are presented in Section~\ref{sec:method}, followed by experimental results in Section~\ref{sec:results}. Finally, we discuss ethical considerations and data-related aspects in the concluding sections.


\section{Related Work}
\label{sec:related}
\paragraph{IWSLT QUE--SPA (2023--2025).}
Since 2023, the International Conference on Spoken Language Translation (IWSLT)\footnote{\url{https://iwslt.org/}} \cite{agostinelli2025findings, ahmad2024findings, agarwal2023findings} has included \emph{Quechua$\to$Spanish} within its Low-Resource/Dialect track. The organizers released a curated ST set of $\sim$1h40m of parallel Quechua speech with \spanish translations and, for unconstrained participation, tens of hours ($\sim$48--60h) of fully transcribed Quechua drawn from Siminchik along with several machine translated (MT) texts. \cite{zevallos2022huqariq, ortega2020neural, ZEVALLOS18.4}

In the past three years, the \emph{Quechua$\to$Spanish} submissions to IWSLT have shown consistent BLEU \cite{papineni2002bleu} score improvements via multilingual pretrained models (e.g., SpeechT5 \cite{ao2022speecht5}, SeamlessM4T \cite{barrault2023seamlessm4t}) and synthetic data, with 2024--2025 findings documenting steady gains and practical recipes for both cascaded and end-to-end ST \cite{e-ortega-etal-2025-quespa, quespa2024, quespa2023, naver_iwslt2023}. These results validate the impact of \emph{domain-aligned} and \emph{synthetic} resources for low-resource speech tasks.

\paragraph{AmericasNLP (2021--2025).}
Alternatively, another workshop dedicated to the preservation of American indigenous languages called AmericasNLP\footnote{\url{https://turing.iimas.unam.mx/americasnlp/}} has consistently included Quechua as a language for MT and ASR tasks since its inception in 2021.

The AmericasNLP shared tasks and findings report measurable gains in Spanish$\leftrightarrow$\quechua MT using pretrained multilingual models and carefully curated corpora, including \emph{legal-domain} content such as constitutions/laws \cite{americasnlp_findings_2023,americasnlp_ehu_2023}. These insights support our choice of the \constitution\ as a high-value, stable source for speech synthesis and evaluation to be used for future tasks in the political arena.

\paragraph{Quechua ASR and augmentation.}
For Quechua ASR, \citet{zevallos22_interspeech} proposed TTS-driven augmentation and reported $\sim$8.7\% absolute word error rate (WER) reduction; \citet{zevallos22_interspeech} corroborated improvements using wav2letter++ \cite{pratap2019wav2letter++} with synthetic data. In broader Indigenous ASR, self-supervised models (e.g., XLS-R \cite{babu2021xls}, mHuBERT \cite{boito2024mhubert}) transfer surprisingly well, indicating that additional labeled pairs from TTS can be highly effective \cite{chen_lrec_2024,romero_arxiv_2024,romero_mdpi_2024}. Their techniques were used in the latest IWSLT team QUESPA submission. \cite{e-ortega-etal-2025-quespa}

\paragraph{TTS frameworks for low-resource.}

Recent open-source TTS ecosystems have lowered the barriers to developing speech synthesis systems for low-resource languages. Coqui\footnote{\url{https://github.com/coqui-ai/TTS}} TTS enables cross-lingual synthesis through \emph{XTTS} \cite{casanova24_interspeech}, supporting multilingual speaker adaptation and facilitating knowledge transfer in unbalanced Quechua--Spanish scenarios. In parallel, diffusion- and flow-matching approaches, such as \emph{F5-TTS} \cite{chen-etal-2025-f5} and \emph{DiFlow-TTS} \cite{nguyen2025diflow}, improve robustness and prosodic modeling, making them well suited for low-resource and noisy data conditions.

\section{Method and Settings}
\label{sec:method}
\begin{figure}[b]
  \centering
  \includegraphics[width=\linewidth]{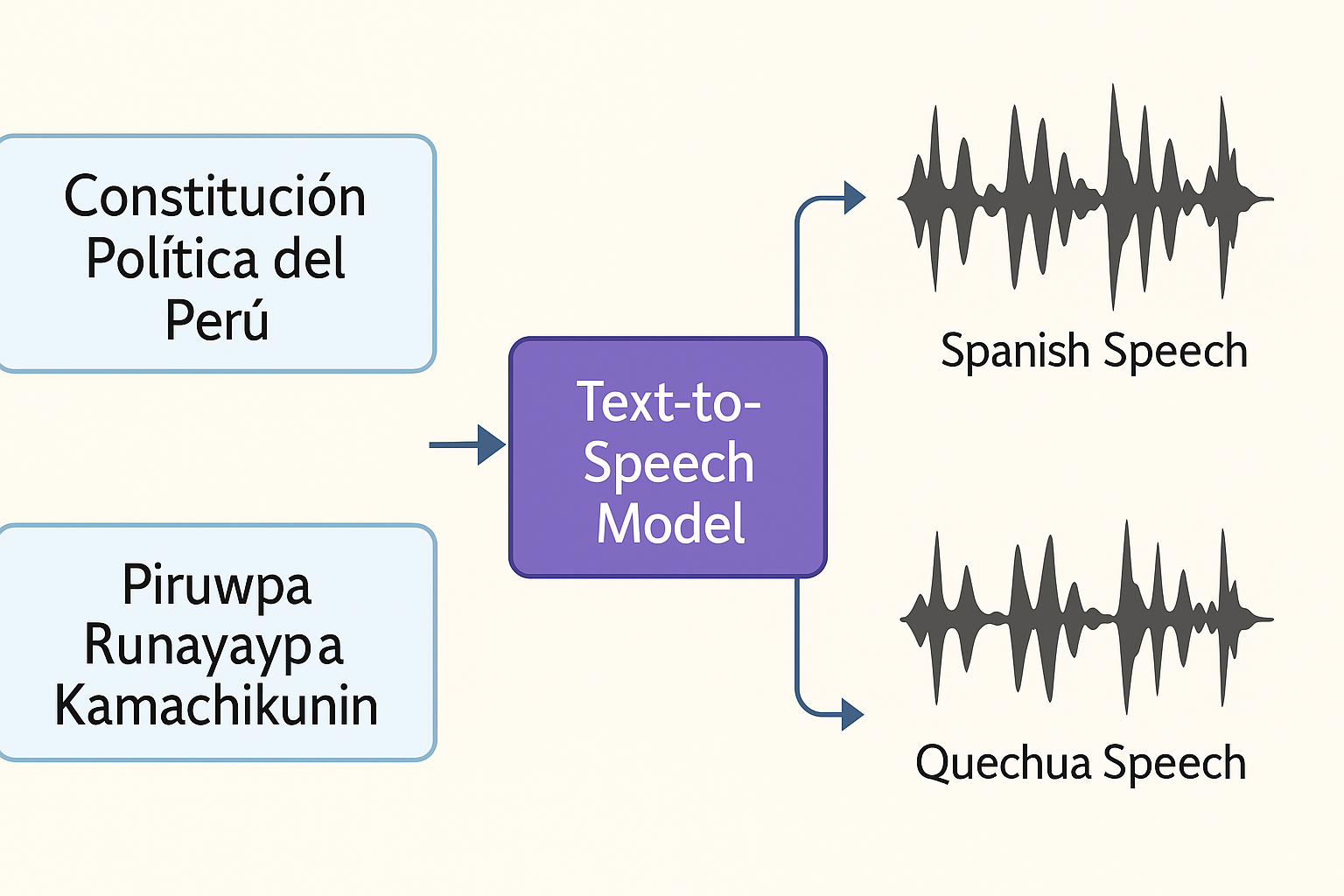}
  \caption{Spanish and Quechua Text to Speech Model.}
  \label{fig:tts_flow}
  \end{figure}

\subsection{Corpus and Normalization}

We distinguish between speech–text corpora used for model training and text-only resources employed exclusively for evaluation. For Quechua, we utilize the Siminchik \cite{ZEVALLOS18.4} and Lurin \cite{zevallos22_interspeech} corpora, which provide approximately 97.5 and 83.3 hours of fully transcribed Southern Quechua speech, respectively. The Lurin corpus comprises approximately 8,000 read-speech sentences collected from lexicographic and literary sources, including the Collao–Spanish and Chanka–Spanish dictionaries. Together, these datasets constitute approximately 180 hours of aligned speech–text data.

Both corpora contain a mixture of long utterances ($\approx 30$s) and very short segments ($\approx 1$s). Since extremely short clips provide limited prosodic information and may negatively affect duration and alignment modeling in neural TTS systems, we apply duration-based filtering and discard all segments shorter than one second and some of low quality (e.g., music, noise, silence). This quality-control step reduces the total amount of training data by approximately 140 hours, resulting in a final curated corpus of approximately 40 hours.

All Quechua transcriptions are normalized using a morphological parser and normalizer to convert surface forms into standardized Southern Quechua representations, thereby reducing orthographic variability and data sparsity in this highly agglutinative language. Additional cleaning procedures include segmentation refinement, removal of misaligned samples, and consistency checks across morphological boundaries, following prior work on low-resource Quechua data preparation.

For Spanish, we leverage a large collection of open-source speech corpora curated for high-quality text-to-speech training \cite{guevara-rukoz-etal-2020-crowdsourcing}. The combined dataset includes professionally recorded and crowdsourced read speech spanning multiple Spanish dialects, such as Peninsular, Argentinian, Chilean, Colombian, Peruvian, Puerto Rican, and Venezuelan varieties, as well as TEDx Spanish recordings. In total, the training set comprises approximately 218 hours of annotated Spanish audio. All audio samples undergo resampling, amplitude normalization, and sentence-level segmentation prior to training to ensure consistency and precise alignment between speech and transcriptions across sources.

In addition to training resources, we compile authoritative \spanish text of the \constitution\ and catalogue \quechua versions/segments referenced by public repositories \cite{constitucion_peru_quechua}. We retain Title/Article headings and segment into sentences to produce \emph{article-aligned} JSON (IDs, language tags). We normalize punctuation, numbers (cardinals/ordinals, dates), legal abbreviations (e.g., \textit{Art.}, \S), and expand references and Roman numerals. For \quechua, we adopt a rule-based grapheme-to-phoneme front end that preserves orthographic conventions in the source and marks morphological hyphenation minimally.

\subsection{Models and Training}

We synthesize audio using three SOTA text‑to‑speech systems, XTTS‑v2\footnote{\url{https://github.com/coqui-ai/TTS}}, F5‑TTS\footnote{\url{https://github.com/SWivid/F5-TTS}}, and DiFlow‑TTS\footnote{\url{https://github.com/Tobertz-max/DiFlow-TTS}}, trained or fine‑tuned following their respective official recipes and optimized on NVIDIA H100 GPUs for efficient convergence and reproducibility. XTTS‑v2 uses a GPT‑based encoder‑decoder stack with standard Coqui\footnote{\url{https://github.com/coqui-ai/TTS}} hyperparameters, AdamW optimization, cosine learning rate scheduling, and FP16 mixed precision; pre-processing follows Coqui dataset tools, and pretrained checkpoints accelerate convergence while enabling multilingual and voice‑conditioned synthesis. F5‑TTS adopts the flow‑matching paradigm, with text processed through ConvNeXt \cite{woo2023convnext} encoders and denoised via the Diffusion Transformer backbone; it preserves default flow‑matching objectives, step schedules, and sampling strategies such as Sway Sampling for inference efficiency. DiFlow‑TTS leverages discrete flow matching over factorized codec tokens for zero‑shot synthesis, maintaining discrete flow losses and factorized prediction heads for prosody and acoustics, following open‑source training procedures. Across all models, learning rates (1e‑4 and 5e‑5), weight decay, gradient clipping, mixed precision, and batch sizing are tuned for stability and generalization; training spans tens to hundreds of thousands of steps with early stopping based on validation loss and audio quality. Generated audio is saved as 22.05–24 kHz WAV files with consistent amplitude normalization, sentence alignment, and metadata to support systematic evaluation and reproducibility.



\subsection{Evaluation}

We adopt an objective and proxy-perceptual evaluation protocol tailored to low-resource TTS settings and fully aligned with the reported results. All metrics are computed on \quechua audio synthesized from constitutional text.

\begin{itemize}[leftmargin=1.25em]
  \item \textbf{Perceptual Quality (Proxy)}: We report \textbf{UTMOS}, an objective estimator of mean opinion scores, to approximate perceived naturalness of synthesized speech without requiring large-scale human evaluations.
  
  \item \textbf{Speaker Consistency}: Cross-utterance speaker stability is measured using \textbf{SIM-O}, computed as the cosine similarity between speaker embeddings extracted from synthesized speech. Higher values indicate more consistent voice characteristics across generated samples.
  
  \item \textbf{ASR-based Intelligibility}: We report Word Error Rate (\textbf{WER}) on synthesized speech using language-specific ASR systems. For Quechua, we employ the \textbf{QUESPA} self-supervised ASR model \cite{e-ortega-etal-2025-quespa}, which is specifically designed for low-resource Andean languages. Both systems are used off-the-shelf without additional fine-tuning.
  
  \item \textbf{Prosodic Accuracy}: To evaluate prosodic stability, we report root mean squared error for fundamental frequency (\textbf{RMSE}$_{F0}$) and energy (\textbf{RMSE}$_E$), computed between reference and synthesized signals. Lower values indicate more accurate modeling of pitch contours and loudness dynamics.
\end{itemize}

This evaluation setup enables consistent comparison across models while remaining feasible in a low-resource bilingual scenario, where large-scale subjective evaluations are often impractical.

\section{Results}
\label{sec:results}

\begin{table*}[]
\centering
\begin{tabular}{l|c|ccccc}
\hline
\textbf{Model} & \textbf{\#Params} & \textbf{UTMOS $\uparrow$} & \textbf{SIM-O} $\uparrow$ & \textbf{WER} $\downarrow$ & $\textbf{RMSE}_\textbf{F0} \downarrow$ & $\textbf{RMSE}_\textbf{E} \downarrow$ \\ \hline
XTTS-V2 & 470M  &  3.22  &   0.53   &    0.19          &    21.03               &        0.021          \\
F5-TTS         & 336M              &    3.23           &        0.60        &      0.19        &       15.17            &     0.017             \\
DiFLOW-TTS     & 164M              &   3.31             &      0.49          &     0.16         &       10.24            &    0.011              \\ \hline
\end{tabular}
\caption{Objective and perceptual evaluation of \quechua synthesized speech using constitutional text as input. All metrics are computed on audio generated from \quechua text of the Constitution, leveraging cross-lingual training with \spanish as a high-resource language. Higher is better for UTMOS and SIM-O, while lower is better for WER, RMSE$_{F0}$, and RMSE$_E$.}

\label{tab:1}
\end{table*}

Table~\ref{tab:1} reports the quantitative evaluation of the three TTS systems considered in this work. In line with our low-resource setting, all models are trained or adapted using a bilingual setup combining \spanish (high-resource) and \quechua (low-resource) data, with the goal of assessing how cross-lingual transfer impacts perceptual quality, intelligibility, and prosodic stability in \quechua synthesis.

Across all metrics, DiFLOW-TTS achieves the most favorable trade-off between model size and synthesis quality. Despite being the smallest model (164M parameters), it obtains the highest UTMOS score (3.31) and the lowest WER (0.16), indicating improved naturalness and intelligibility under limited \quechua supervision. Crucially, it also yields the lowest $\text{RMSE}_\text{F0}$ and $\text{RMSE}_\text{E}$, suggesting that cross-lingual acoustic patterns learned from \spanish are transferred more effectively to stabilize prosody in \quechua.

F5-TTS, with 336M parameters, shows competitive perceptual quality (UTMOS 3.23) and achieves the highest SIM-O score (0.60), reflecting stronger speaker consistency across languages. While its WER matches that of XTTS-V2, its substantially lower prosodic errors point to a more robust exploitation of shared phonetic and rhythmic structure between \spanish and \quechua, even when trained with limited target-language data.

XTTS-V2, the largest model evaluated (470M parameters), provides stable baseline performance in the cross-lingual scenario but does not fully capitalize on parameter scale. Its comparatively higher F0 and energy reconstruction errors indicate less precise prosodic transfer, highlighting that larger models do not necessarily yield better outcomes in low-resource bilingual settings.

Overall, these results underline the importance of cross-lingual learning over model scaling for low-resource TTS. Leveraging \spanish data enables all systems to synthesize intelligible and natural \quechua speech; however, architectures explicitly designed to control temporal and prosodic dynamics exhibit superior transfer efficiency. This makes such models particularly suitable for low-resource languages, where data scarcity and deployment constraints are central considerations.





\section{Ethics \& Limitations}
We avoid any personally identifiable or celebrity-like voices; voices are synthetic or consented. We document dialect scope (e.g., Cusco Collao influence), orthographic choices, and intended use. We encourage Indigenous data governance practices and feedback from \quechua media/community groups\footnote{\url{https://latamjournalismreview.org}}. The TTS is not a substitute for professional legal interpretation and may require style tuning for ceremonial or court settings.

\section{Conclusion}
In this work, we set out to make the Peruvian Constitution accessible in \quechua through high-quality synthesized speech, addressing a concrete gap at the intersection of language rights and speech technology. Given the severe data scarcity of \quechua, we adopt a bilingual training strategy that leverages \spanish as a high-resource language to enable effective cross-lingual transfer for TTS.

Our results show that cross-lingual TTS models trained on \spanish and \quechua data can generate intelligible and natural \quechua speech from legal-domain text, with competitive perceptual quality and stable prosody. Notably, model architectures that explicitly control temporal and prosodic dynamics achieve better performance despite substantially fewer parameters, underscoring that architectural design and data alignment are more critical than scale in low-resource settings.

Beyond the quantitative evaluation, the released resources, including bilingual aligned text, reproducible inference pipelines, and synthesized audio of the entire Constitution; constitute a practical and reusable asset for the community. This bilingual legal TTS resource provides immediate value for \quechua accessibility while also serving as a domain-matched seed for downstream ASR and speech-to-text research, where synthetic data from trusted sources has been shown to yield measurable benefits.

Overall, this work demonstrates that high-impact speech resources for low-resource languages can be built using open tools and cross-lingual learning, even under limited supervision. We hope that this effort encourages further research on legally grounded, socially relevant speech technologies for indigenous and underrepresented languages.

\section{Data and Prompt Availability}
\label{se:appendix:prompts}
Due to the appendix constraint and for anonymity, we omit the data and prompts used. We will deliver them upon positive acceptance.

\section*{Acknowledgements}
We thank the speaker communities and language workers associated with the Quechua work performed. We also thank the maintainers and staff of documentation archives and repositories consulted in this study for curating and providing access to materials and metadata. Finally, we thank the Political NLP 2026 reviewers for constructive feedback that improved the camera-ready version.

\bibliographystyle{lrec2026-natbib}
\bibliography{references}

\end{document}